\documentclass{IET}%%%%where IET is the template name

\newcommand{\argmin}{\arg\!\min}

\begin{document}

\title{Palmprint Recognition Using Deep Scattering Convolutional Network}

\author[1.*]{Shrevin Minaee}
\affil{New York University}

\author[1]{Yao Wang}
%%%% By default, the citations will come automatically,
%%%% The optional bracket "[2.*]" is used  to display the corresponding author symbol
%\affil{New York University}

%%%% Corresponding author detail must placed here
\affil[*]{shervin.minaee@nyu.edu}

%Due to the rich set of features existing in palmprints and the simplicity of palmprint image acquisition, it is considered one of the most popular biometrics.
\abstract{Palmprint recognition has drawn a lot of attention during the recent years. 
Many algorithms have been proposed for palmprint recognition in the past, majority of them being based on features extracted from the transform domain.
Many of these transform domain features are not translation or rotation invariant, and therefore a great deal of preprocessing is needed to align the images.
In this paper, a powerful image representation, called scattering network/transform, is used for palmprint recognition. Scattering network is a convolutional network where its architecture and filters are predefined wavelet transforms.
The first layer of scattering network captures similar features to SIFT descriptors and the higher-layer features capture higher-frequency content of the signal which are lost in SIFT and other similar descriptors. 
After extraction of the scattering features, their dimensionality is reduced by applying principal component analysis (PCA) which reduces the computational complexity of the recognition task. Two different classifiers are used for recognition: multi-class SVM and minimum-distance classifier.
The proposed scheme has been tested on a well-known palmprint database and achieved accuracy rate of 99.95\% and 100\% using minimum distance classifier and SVM respectively.}

\maketitle

\section{Introduction}
Identification has always been required in critical tasks and applications. Throughout history, there were always attempts to make this process flawless and secure, mostly to prevent forgeries. For many centuries, identity was confirmed through an item or a mark.	Nowadays, there are many ways for identification of a person, including passwords, keys and something that is very difficult to duplicate quickly; features of the person himself, also known as biometric data. Biometric-based identification began in the late 19th century with the collection of fingerprints for forensic purposes due to them being unique to every person from whom they are sampled \cite{fing}. Afterwards many other biometrics have been proposed for identity authentication such as palmprints \cite{palm}, iris patterns \cite{iris} and face \cite{face}.
In the more recent works, multimodal biometric systems have also drawn a lot of attention where several biometrics are combined together to achieve more reliable identification.

In this paper we have focused on palmprint recognition. Palmprints are very economical in the sense of acquisition. They can be easily obtained using CCD cameras. They also work in different conditions of weather and are typically time-independent. However, due to sampling limitations, lighting and other factors, they may pose problems like insufficient data due to unclear wrinkles or confusion due to poor image quality. 
%This is the reason there are usually many different samples from every person in the database.
The primary characteristics of a palmprint are the principal lines (including head line, heart line and life line) and wrinkles running through it. The depths and patterns of these lines have a great amount information about the identity of the person. One could extract these lines and use them for identification, but usually, due to the low quality and contrast of palmprint images it is not simple to accurately extract these lines. Therefore it is not reliable to design a system only based on the line information. Subsequently many algorithms have been proposed for palmprint recognition based on different features extracted from palm, including geometric features, wavelet and Fourier features, etc.
In \cite{wave}, Minaee proposed a wavelet-based approach for multispectral palmprint recognition. 
In \cite{dct} Dale proposed a texture-based palmprint recognition using DCT features.
In \cite{wave+dct}, an algorithm is proposed for palmprint recognition which combined both wavelet and DCT features.
Connie proposed a palmprint recognition framework based on principal component analysis and independent component analysis \cite{connie}.
There are also several line-based approaches for palmprint recognition. 
In \cite{minituae}, Cappelli proposed a recognition algorithm based on minutiae.
%Chen \cite{10} proposed a recognition algorithm that primarily uses creases. They extract all creases from a palm and use them for palmprint matching. The main advantage of this algorithm is that it is rotation- and translation-invariant. 
In addition, there have been some works using image coding methods for palmprint recognition, such as Palm Code, Fusion Code, Competitive Code, Ordinal Code \cite{competetive}, \cite{orieant}. 
A survey of palmprint recognition algorithms before 2009 is provided by Kong in \cite{palm}. 
It is worth mentioning that foreground segmentation techniques \cite{seg1}-\cite{seg3} can also be used to separate texture part of the palmprint for more robust feature extraction.
Among more recent works, in \cite{HOL} Jia proposed new descriptor called histogram of oriented line (HOL) inspired by histogram of oriented gradients (HOG) for palmprint recognition and achieved a high recognition rate. In \cite{texture}, Minaee proposed an algorithm based on textural features derived from the co-occurrence matrix.
In \cite{QPCA}, Xu proposed a quaternion principal component analysis approach for multispectral palmprint recognition.

Many of the above algorithm require many pre-processing steps which are specifically designed for a dataset and the final performance largely depends on the quality of those pre-processing steps. 
To reduce this dependency on the dataset, there has been a lot of effort in machine learning and computer vision communities to design supervised and unsupervised feature-learning algorithms which work pretty well over various datasets and problems. These image representations should provide a degree of invariance with respect to certain transformations to make the image classification robust. Along this goal, deep neural networks and dictionary learning algorithms trained on large databases have provided promising result on many benchmarks. For image classification, convolutional neural networks specifically achieved state-of-the-art results on many of the challenging object recognition competitions \cite{Alex}.
In another recent work, a new multi-layer signal representation known as scattering transform/network has been proposed which is a special kind of convolutional network in which the filters and architectures are predefined wavelet filters \cite{scat1}, \cite{scat2}. The scattering network also provides a very rich and discriminative representation of the images.
Due to the tremendous success of deep scattering networks on several image and audio classification benchmarks \cite{scat1}-\cite{scat3} it is interesting to know how this representation works for biometric recognition. 
In the past, we explored the application of scattering network for iris and fingerprint recognition tasks \cite{myiris}, \cite{myfing}. 
In this work, the scattering transformation has been used for palmprint recognition. Scattering transform is applied on each palm image, and features are extracted from the transformed images from different layers of the scattering network.
After that the feature dimensionality is reduced using PCA \cite{PCA}. In the end, multi-class SVM \cite{SVM} and nearest neighbor algorithms are used to perform classification using PCA features. This algorithm has been tested on the well-known PolyU palmprint database \cite{polyu} and achieved very high accuracy rate.

The rest of this paper is organized as follows. Section II describes the features which are used in this work. The detail of scattering features is provided in Section II.A and the PCA algorithm is explained in Section II.B. Section III provides the description of the classification schemes. The minimum distance classifier and support vector machines are explained in sections III.A and III.B respectively. The experimental results and comparisons with other works are provided in Section IV and the paper is concluded in Section V.

\section{Features}
Feature extraction is one of the most important steps in many classification and recognition problems. Usually the more informative and discriminative features we have, the higher accuracy we get. Therefore, many researchers have focused on designing useful features for many specific tasks including object recognition, audio classification, biometrics recognition, etc.
One main difficulty of image classification problems is that images of the same object could have variablity due to translation, scale, rotation, illumination changes, etc. These changes in the images of a single object class are called intra-class variations. 
Depending on the application, it is often desired to extract features that are invariant to some of these transformations.
Various image descriptors have been proposed during past 20 years. SIFT \cite{sift} and HOG \cite{hog} are two popular hand-crafted image descriptors which achieved very good results on several object recognition tasks. These features try to extract the oriented gradient information of the image. 
%Another widely used scheme is the bag-of-visual-words model inspired by document classification where each image is represented as histogram of visual words.
Unfortunately some of these hand-crafted features are not very successful for some of the more challenging object recognition datasets with many object classes and large intra-class variations.
In the more recent works, unsupervised feature learning and deep neural networks achieved state-of-the-art results on various datasets, most notably the Alex-net \cite{Alex} which is trained on ImageNet competition. These algorithms try to learn the features by training a model on a large dataset, or recover the lost information in hand-crafted features.
%Surprisingly, people found that the Alex-net trained on ImageNet also performs well on other datasets like Caltech \cite{alex_generality}. This shows that these deep networks are learning some general image representations which do not depend on the dataset.
Since then, convolutional networks \cite{CNN} and deep learning algorithms have been widely used in many object recognition problems.

In a more recent work, Mallat proposed a multi-layer representation, called scattering network/transform, which is similar to deep convolutional network, but instead of learning the filters and representation, it uses predefined wavelets \cite{scat2}. 
This algorithm has been successfully applied to various object recognition and audio classification problems and achieved state-of-the-art results \cite{scat1}, \cite{scat3}.
In this paper, the application of scattering transform for palmprint recognition is explored. 
Details of scattering transform and its derivation are presented in the next section.

\subsection{Scattering Features}
The scattering network is a deep convolutional network which uses wavelet transform as its filter and can be designed such that it is invariant to transformations such as translation, rotation, etc \cite{scat4}. It provides a multi-layer representation of the signal where, at each layer, the local descriptors of the input signal are computed with a cascade of three operations: wavelet decompositions, complex modulus and a local averaging.
Figure 1 displays an illustration of scattering network applied up to 3 layers.
\begin{figure}[4 h]
\begin{center}
    \includegraphics [scale=0.45] {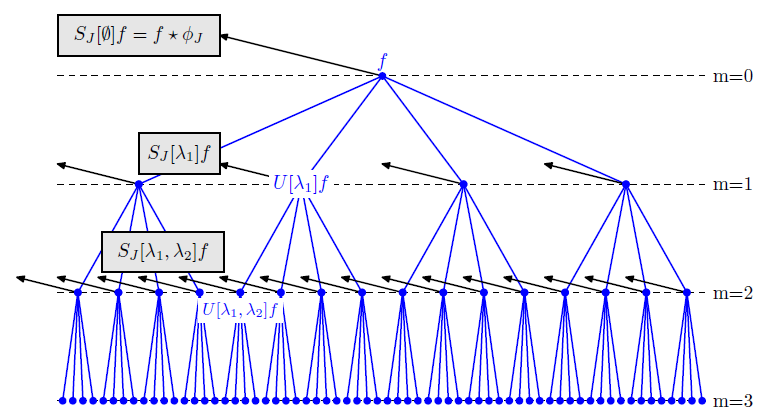}
\end{center}
  \vspace{-0.2cm}
  \caption{Scattering operator applied up to three layers. The output signals in each layer are used as the input for the scattering transformation in the next layer of scattering network \cite{scat2}. The scattering operator at each layer is simply the cascade of three operations: wavelet decompositions, complex modulus (phase removal) and a local averaging.}
\end{figure}

In many object recognition tasks, locally translation-invariant features are preferred, since they provide a robust representation of images. SIFT is one of such descriptors.
As discussed in \cite{scat4}, some of the image descriptors such as SIFT can be obtained by averaging the amplitude of wavelet coefficients, calculated with directional wavelets. This averaging provides some level of local translation invariance, but it also reduces the high-frequency information. Scattering transform is designed such that it recovers the high-frequency information lost by averaging. 
It can be shown that the images in the first layer of the scattering transform are similar to the SIFT descriptor and the coefficients in the higher layers contain higher-frequency information of the image which is lost in SIFT.
Therefore scattering transform not only provides locally translation-invariant features but it also recovers high-frequency information lost in SIFT, resulting in richer descriptors for complex structures such as corners and multiscale texture variations. 
In this work, translation-invariant scattering transform is used and a brief description thereof is provided here.

Suppose we have a signal $f(x)$. The first scattering coefficient is the average of the signal and can be obtained by convolving the signal with an averaging filter $\phi_J$ as $S_{0,J}(f)= f*\phi_J$. The scattering coefficients of the first layer can be obtained by applying wavelet transforms at different scales and orientations, and removing the complex phase and taking their average using $\phi_J$ as shown below:
\begin{gather}
S_{1,J}(f)= |f* \psi_{j_1,\lambda_1}|*\phi_J
\end{gather}
where $j_1$ and $\lambda_1$ denote different scales and orientations. Taking the magnitude of the wavelet coefficients can be thought of as the non-linear pooling functions which are used in convolutional neural networks.
Note that by removing the complex phase of the wavelet coefficients we can make them invariant to local translation, but some of the high-frequency content of the signal will also be lost due to the averaging.

Now to recover the high-frequency contents of the signal, which are eliminated from the wavelet coefficients of first layer, we can convolve  $|f* \psi_{j_1,\lambda_1}|$ by another set of wavelets at scale $j_2<J$, take the absolute value of wavelet coefficients followed by the averaging as:
\begin{gather}
S_{2,J}(f)= ||f* \psi_{j_1,\lambda_1}|*\psi_{j_2,\lambda_2}|*\phi_J
\end{gather}
It can be shown that $|f* \psi_{j_1,\lambda_1}|*\psi_{j_2,\lambda_2}$ is negligible for scales where $2^{j_1} \leq  2^{j_2}$. Therefore we only need to calculate the coefficients for $j_1 >j_2 $.

%The convolution with $\phi_J$ at the second layer removes high frequencies and results in locally translation invariant second-order coefficients. This high-frequency information can be restored again by finer scale wavelet coefficients in the next layers. 
We can continue this procedure to obtain the coefficients of the $k$-{th} layer of the scattering network as:
\begin{gather}
\underset{\ \ \ \ \ \ \ \ \ \ \ \ \ \ \ \ \ \ \ \ \ \ \ \ \ \ j_k<...<j_2<j_1<J, \ (\lambda_1,...,\lambda_k) \in \Gamma^k  }{S_{k,J}(f(x))= ||f* \psi_{j_1,\lambda_1}|*...*\psi_{j_k,\lambda_k}|*\phi_J}
\end{gather}

It is easy to show that the scattering vector of the $k$-{th} layer has a size of $p^k {J \choose k}$ where $p$ denotes the number of different orientations and $J$ denotes the number of scales.
A scattering vector is formed as the concatenation of the coefficients of all layers up to $m$ which has a size of $\sum_{k=0}^m{p^k {J \choose k}}$. For many signal processing applications, a scattering network with two or three layers will be enough.

The transformed images of the first and second layers of the scattering transform for a sample palmprint are shown in Figures 2 and 3. These images are derived by applying bank of filters of 5 different scales and 6 orientations. As it can be seen, each transformed image in the first layer is capturing the information along one direction.

\begin{figure}[2 h]
\begin{center}
    \includegraphics [scale=0.45] {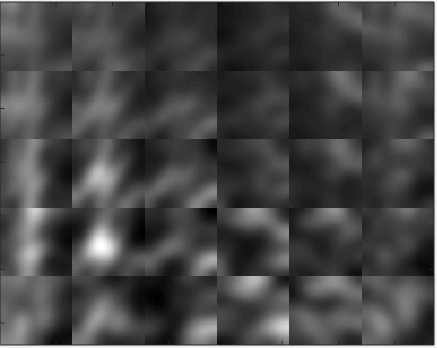}
\end{center}
  \vspace{-0.2cm}
  \caption{The images from the first layer of the scattering transform}
\end{figure}
\begin{figure}[3 h]
\begin{center}
    \includegraphics [scale=0.5] {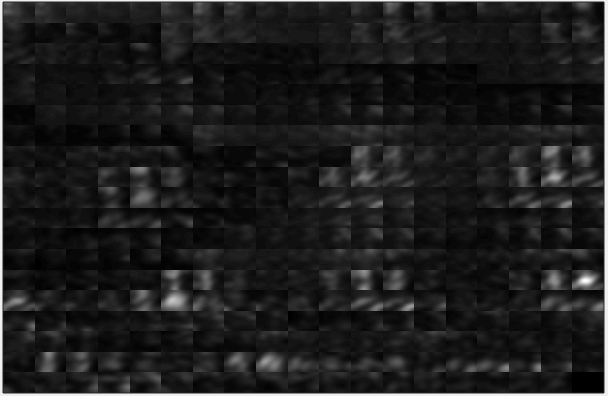}
\end{center}
\vspace{-0.2cm}
  \caption{The images from the second layer of the scattering transform}
\end{figure}

In the end, we extract the mean and variance of each scattering transformed image to form the scattering feature vector of length $2\sum_{k=0}^m{p^k {J \choose k}}$. One can also extract further information from each transformed image to form the scattering feature vector. Before using these features for recognition, their dimensionality is reduced using principal component analysis (PCA). A brief description of PCA is provided in the following section.

\subsection{Principal Component Analysis}
In many applications, one needs to reduce the dimensionality of the data to make the algorithm faster and more efficient.
There have been many dimensionality reduction algorithms proposed in the past. Principal component analysis (PCA) is a powerful dimensionality reduction algorithm used in many areas including computer vision, document analysis, biology, statistic \cite{PCA}. EigenFace \cite{eigenface} is one of the most notable applications of PCA in computer vision where each face is projected onto principal faces and used for face recognition. Given a set of correlated variables, PCA transforms them into another domain such that the transformed variables are linearly uncorrelated. This set of linearly uncorrelated variables are called principal components. PCA is usually defined in a way that the first principal component has the largest variance, and the second one has the second largest variance and so on. After extracting the principal components, we can reduce the dimensionality of the data by projecting them onto a subset of principal components with the largest variance.
%PCA has a lot of applications in computer vision. Eigenface is one representative application of PCA in computer vision, where PCA is used for face recognition.

Let us assume we have a dataset of $M$ images and $\{f_1,f_2,...,f_M\}$ denotes their feature vectors where $f_i \in \mathbb{R}^d$. To apply PCA, all features need to be centered first by removing their mean:  $z_i= f_i- \bar{f}$ where $\bar{f}= \frac{1}{M} \sum_{i=1}^M f_i$.
Then the covariance matrix of the centered images is calculated as:
\begin{gather*}
C= \sum_{i=1}^M z_i z_i^T 
\end{gather*}
Next the eigenvalues $\lambda_j$ and eigenvectors $\nu_j$ of the covariance matrix $C$ are computed. Suppose $\lambda_j$'s are ordered based on their values.
The dimensionality of the data can be reduced by projecting them on the first $K (\ll d)$ principal vectors as:
\begin{gather*}
\hat{z_i}= (\nu_1^{T} z_i, \nu_2^{T} z_i,..., \nu_K^{T} z_i)
\end{gather*}

By keeping $k$ principal components, the percentage of retained variance can be found as $\frac{\sum_{j=1}^k \lambda_j}{\sum_{j=1}^d \lambda_j }$. 
%One issue is how to choose the value $k$, number of principal components. One simple way to choose $k$ would be to pick a value such that the above ratio is less than $\epsilon$, where $\epsilon$ is usually chosen between 95\% to 99\%.

\section{Recognition Algorithm}
After extracting the features of all images in the dataset, a classifier should be trained to find the class label of the samples. There are various classifiers which can be used for this task including support vector machine (SVM), neural network and minimum distance classifier (also known as nearest neighbor). In this work two classifiers are used to for recognition: minimum distance classifier and support vector machine \cite{SVM} which are quite popular in image classification area.
A brief overview of minimum distance classifier and SVM are presented in the next sections. 

\subsection{Minimum Distance Classifier}
Minimum distance classifier is a popular algorithm in the biometric recognition and template matching area. To find the class label of a test sample using the minimum distance classifier, we basically need to find the distances between the features of the test sample and those of the training samples, and pick the label of the class with the minimum distance to the unknown as the predicted label.
Therefore if $f_{(t)}$ denotes the features of a test sample and $f_{k}$ denotes the features of the $k$-th sample in our dataset, minimum distance assigns the test sample to one of the samples in the dataset such that:
\begin{gather}
k^*=\argmin_k \big[ dis(f_{(t)},  f_{k}) \big]
\end{gather}
Here, Euclidean distance is used as our metric.

\subsection{Support Vector Machine}
Support vector machine is a well-known machine learning approach designed for classification. It has been widely used in computer vision applications during the past decades. Due to the convexity of the cost function in SVM, it is guaranteed to find the global  of model parameters.
Let us assume we want to separate the set of training data $(x_1,y_1)$, $(x_2,y_2)$, ..., $(x_n,y_n)$ into two classes where 
$x_i \in R^d $ is the feature vector and $y_i \in \{-1,+1\}$ is the class label. Let us assume two classes are linearly separable with a hyperplane $w.x+b=0$. With no other prior knowledge about the data, the optimal hyperplane can be defined as the one with the maximum margin (which results in the minimum expected generalization error). SVM finds this hyperplane such that $w.x_i+b \geq 1$ for $x_i$ in the first class and $w.x_i+b \leq -1$ for $x_i$ in the second class, and maximizes the distance between these two which is $\frac{2}{||w||_2}$. This problem can be written as:
\begin{equation}
\begin{aligned}
& \underset{w,b}{\text{maximize}}
& & \frac{2}{||w||_2}  \\
& \text{subject to}
& & y_i(w.x_i+b) \geq 1, \; i = 1, \ldots, n.
\end{aligned}
\end{equation}
We can convert this problem into the following minimization problem:
\begin{equation}
\begin{aligned}
& \underset{w,b}{\text{minimize}}
& & \frac{1}{2} ||w||^2 \\
& \text{subject to}
& & y_i(w.x_i+b) \geq 1, \; i = 1, \ldots, n.
\end{aligned}
\end{equation} \\
%\begin{equation}
%L(w,b,\alpha)= \frac{1}{2} ||w||^2+ \sum_{i=1}^{n} \alpha_i \big[ y_i(b- w.x_i) -1 \big]
%\end{equation}
Since this problem is convex, we can solve it by looking at the dual problem and introducing Lagrange multipliers $\alpha_i$
which results in the following classifier:
\begin{equation}
f(x)= \mathbf{sign}(\sum_{i=1}^{n} \alpha_i y_i w.x+b)
\end{equation}
where $\alpha_i$, $w$ and $b$ are calculated by SVM learning algorithm ($w$ has a closed-form solution). Interestingly, after solving the dual optimization problem, most of the $\alpha_i$'s are zero; those data points $x_i$ which have nonzero $\alpha_i$ are called support vectors.
There is also a soft-margin version of SVM which allows for mislabeled examples. If there exists no hyperplane that can split the ``-1'' and ``+1'' examples, the soft-margin method will choose a hyperplane that splits the examples as cleanly as possible, while still maximizing the distance to the nearest cleanly split examples \cite{SVM}. It introduces a penalty term in the primal optimization problem with misclassification penalty of $C$ times the degree of misclassification.

To derive a nonlinear classifier, one can map the data from the input space into a higher-dimensional feature space $\mathcal{H}$ as: $x\rightarrow \phi(x)$, so that the classes are linearly separable in the feature space \cite{kernel_SVM}. If we assume there exists a kernel function where $k(x,y)= \phi(x).\phi(y)$, then we can use the kernel trick to construct nonlinear SVM by replacing all inner products $x.y$ with $k(x,y)$ which results in the following classifier:
\begin{equation}
f_n(x)= \mathbf{sign}(\sum_{i=1}^{n} \alpha_i y_i K(x,x_i)+b)
\end{equation}

To derive multi-class SVM for a set of data with $M$ classes, we can train $M$ binary classifiers which can discriminate each class against all other classes, and choose the class which classifies the test sample with the greatest margin (one-vs-one). 
In another approach, we can train  a set of $M \choose 2$ binary classifiers, each of which separates one class from another and  choose the class that is selected by the most classifiers. There are also other ways to do multi-class classification using SVM.
For further details and extensions to multi-class settings, we refer the reader to \cite{multi_SVM}.

\section{Experimental Results}
A detailed description of experimental results is provided in this section.
First, let us describe the parameter values of our algorithm and the database which is used for experimental study.

We have tested the proposed algorithm on the PolyU palmprint database \cite{polyu} which has 6000 palmprints sampled from 500 persons. Each palmprint is taken under 4 different lights in two different days, is preprocessed and has a size of 128x128. %We have only used the images in the blue spectrum for our experiment.

Each image is first divided into blocks of size 32x32. Then for each block the scattering transform is applied up to two levels with a set of filter banks with 5 scales and 6 orientations, resulting in 391 transformed blocks. From each transformed-block, the mean and variance are calculated and used as features, resulting in 12512 scattering features for the entire image. For the scattering transformation, we used the software implemented by Mallat's group \cite{scat}. 
The corresponding wavelet filter banks with 5 scales and 6 orientations, which are used in this paper, are shown in Figure 4.
\begin{figure}[4 h]
\begin{center}
    \includegraphics [scale=0.35] {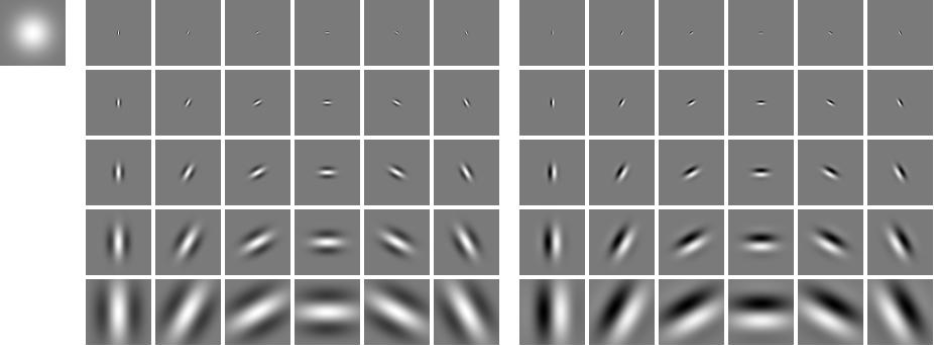}
\end{center}
  \vspace{-0.2cm}
  \caption{The real and imaginary parts of filters used here \cite{scat}. The top left image corresponds to $\phi$, the averaging filter. The left and right halves of images correspond to the real and imaginary parts of $\psi_{j,\lambda}$ respectively, which are arranged according to scales (rows) and orientations (columns).}
\end{figure}

Then PCA is applied to all features and the first K PCA features are used for recognition. Multi-class SVM and minimum distance classifier are used for the template matching. For SVM, we have used the LIBSVM library \cite{libsvm}, with a linear kernel and the penalty cost of $C=1$.

In the first experiment, we explored the effect of using different numbers of features on the recognition accuracy where SVM is used for classification. We have used half of the images in the dataset for training and the other half for testing. Figure 5 shows the recognition accuracy for different numbers of PCA features. As we can see, using 200 PCA features, we are able to get 100\% accuracy rate.
\begin{figure}[4 h]
\begin{center}
    \includegraphics [scale=0.48] {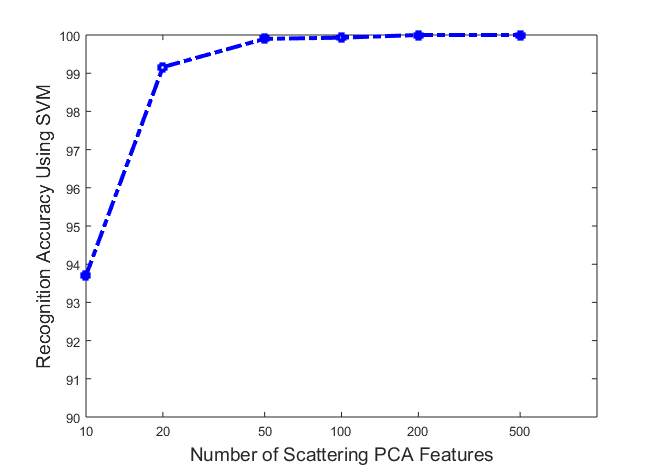}
\end{center}
\vspace{-0.2cm}
  \caption{Recognition accuracy using multi-class SVM}
\end{figure}

For the next experiment, we explored recognition accuracy of using different numbers of PCA features where minimum distance classifier is used for recognition. 
Figure 6 shows the recognition accuracy using minimum distance classifier.
As we can see, the recognition accuracy of minimum distance classifier is slightly less than SVM, but it is still very high.
We can see by using 100 PCA features, we get an accuracy of more than 98\%, and by using 700 PCA features, we get an accuracy of 99.43\%.
\begin{figure}[5 h]
\begin{center}
    \includegraphics [scale=0.58] {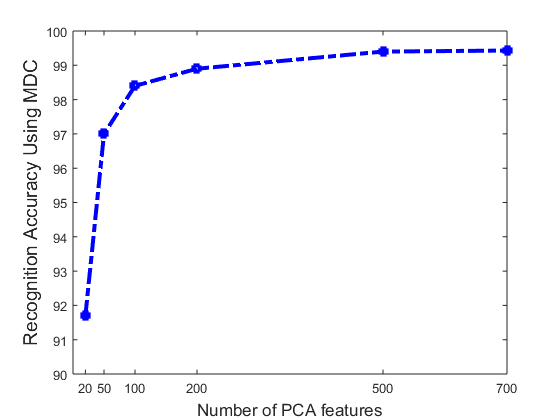}
\end{center}
\vspace{-0.2cm}
  \caption{Recognition accuracy by using minimum distance classifier}
\end{figure}

In another experiment, we explored how the recognition accuracy changes by using different numbers of training samples.
Each time we get $k$ samples from each person as the training and test the algorithm on the remaining $12-k$ samples. The result of this experiment for SVM and minimum distance classifier are shown in Figure 7 and Figure 8 respectively.
\begin{figure}[6 h]
\begin{center}
    \includegraphics [scale=0.48] {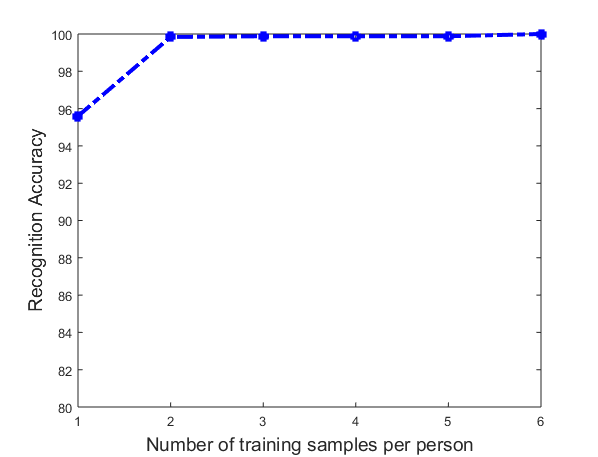}
\end{center}
\vspace{-0.2cm}
  \caption{Recognition accuracy by using different numbers of training samples using SVM}
\end{figure}

\begin{figure}[6 h]
\begin{center}
    \includegraphics [scale=0.48] {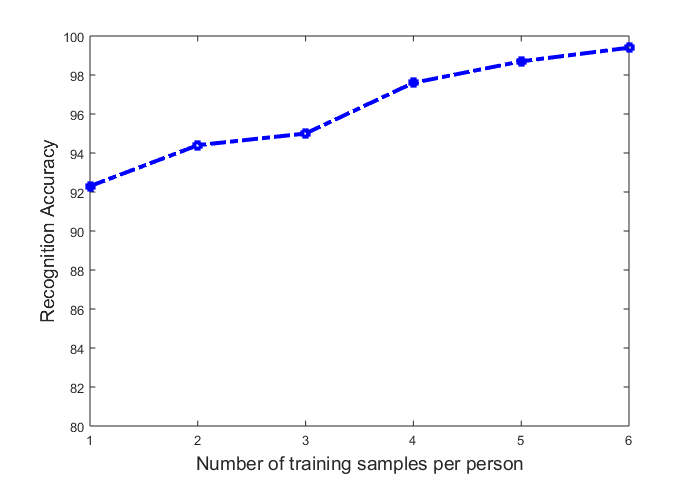}
\end{center}
\vspace{-0.2cm}
  \caption{Recognition accuracy by using different numbers of training samples using minimum distance classifier}
\end{figure}
For SVM, even by using 2 samples as training for each person, we are able to get an accuracy of 99.84\%, and for minimum distance classifier we will get an accuracy of 94.4\% which is quite promising. This shows the power of scattering features which can capture discriminative features even in the case of limited training data.

The comparison of the accuracy of the proposed algorithm with several other recent works is presented in Table 1.
Here KPCA+GWR means that Gabor wavelet is used and kernel-PCA is applied on the features for recognition. Gabor based RCM shows the region covariance matrix derived from the Gabor wavelet. 

\begin{table} [h]
\centering
  \caption{A comparison of recognition accuracy of the proposed scheme and previous approaches }
  \centering
\begin{tabular}{|m{8cm}|m{2.5cm}|}
\hline
\ \ \ \ \ \ \ \ \ \ \ \ \ \ \ \ \ \ \ \ Method & Recognition \ \ \ \  Accuracy\\
\hline
Local binary pattern (LBP)  & \ \ \ 82.3\% \\
\hline
EigenPalm (EP) \cite{GBR}  & \ \ \  91.25\% \\
\hline
FisherPalm (FP) \cite{GBR} & \ \ \   92.32\% \\
\hline
Image level fusion by PCA \cite{QPCA}  & \ \ \  95.17\% \\
%\hline
%KPCA+GWR \cite{GBR} & \ \ \  95.17\% \\
\hline
Gabor-based RCM (GRCM) \cite{GBR} & \ \ \  96\% \\
\hline
Enhanced GRCM (EGRCM) \cite{GBR} & \ \ \ 98\% \\
\hline
Quaternion PCA \cite{QPCA} & \ \ \ 98.13\% \\
\hline
PCA on HOG \cite{HOL} & \ \ \ 98.73\% \\
\hline
KPCA on HOG \cite{HOL} & \ \ \ 98.77\% \\
\hline
Quaternion PCA+Quaternion DWT \cite{QPCA} & \ \ \ 98.83\% \\
\hline
LDA on HOG \cite{HOL} & \ \ \ 99.17\% \\
\hline
KLDA on HOG \cite{HOL} & \ \ \ 99.17\% \\
\hline
The proposed scheme using KNN & \ \ \ 99.4\% \\
\hline
PCA on HOL \cite{HOL} & \ \ \ 99.73\% \\
\hline
KPCA on HOL \cite{HOL} & \ \ \ 99.73\% \\
\hline
The proposed scheme using SVM & \ \ \ 100\% \\
\hline
\end{tabular}
\label{TblComp}
\end{table}

As it can be seen, the proposed algorithm achieves higher accuracy than previous algorithms on this dataset. The main reason is that the scattering features are able to extract a lot of high-frequency contents of the signal which provide a great amount of discriminative information.

%We have also tested the performance of this algorithm for verification task. 

The experiments are performed using MATLAB 2015 on a laptop with Core i5 CPU running at 2.6GHz. 
It takes around 0.09 second for each image to perform template matching using the proposed algorithm. 
Therefore it is fast enough to be implemented in electronic devices for real-time applications using energy-efficient algorithms \cite{eragh1}.

\section{Conclusion}
This paper proposed an algorithm for palmprint recognition using convolutional scattering network. These scattering features are locally invariant and carry a lot of high-frequency information which are lost in other descriptors such as SIFT. 
Then PCA is applied on the features to reduce their dimensionality. In the end, multi-class SVM and nearest neighbor classifier are used to perform template matching.
We believe the multi-layer representation provided by scattering network provides great discriminating power which can benefit biometrics recognition tasks.
In the future, we will investigate to further explore the application of the scattering network for other image classification problems.

\section*{Acknowledgment}
The authors would like to thank Stephane Mallat's research group at ENS for providing the software implementation of the scattering transform. We would also like to thank the CSIE group at NTU for providing the LIBSVM software, and biometric research group at PolyU Hong Kong for providing the palmprint dataset.

\end{document}